# DLGA-PDE: Discovery of PDEs with incomplete candidate library via combination of deep learning and genetic algorithm

Hao Xu[a], Haibin Chang[a,*], and Dongxiao Zhang[b,c,*]

[a] BIC-ESAT, ERE, and SKLTCS, College of Engineering, Peking University, Beijing 100871, P. R. China

[b] School of Environmental Science and Engineering, Southern University of Science and Technology, Shenzhen 518055, P. R. China

[c] Intelligent Energy Lab, Frontier Research Center, Peng Cheng Laboratory, Shenzhen 518000, P. R. China

[*] Corresponding author.

E-mail address: 390260267@pku.edu.cn (H. Xu); changhaibin@pku.edu.cn (H. Chang); donzhang01@gmail.com (D. Zhang).

**Abstract**: Data-driven methods have recently been developed to discover underlying partial differential equations (PDEs) of physical problems. However, for these methods, a complete candidate library of potential terms in a PDE are usually required. To overcome this limitation, we propose a novel framework combining deep learning and genetic algorithm, called DLGA-PDE, for discovering PDEs. In the proposed framework, a deep neural network that is trained with available data of a physical problem is utilized to generate meta-data and calculate derivatives, and the genetic algorithm is then employed to discover the underlying PDE. Owing to the merits of the genetic algorithm, such as mutation and crossover, DLGA-PDE can work with an incomplete candidate library. The proposed DLGA-PDE is tested for discovery of the Korteweg–de Vries (KdV) equation, the Burgers equation, the wave equation, and the Chaffee-Infante equation, respectively, for proof-of-concept. Satisfactory results are obtained without the need for a complete candidate library, even in the presence of noisy and limited data.

## 1. Introduction

With the rapid development of data processing capabilities of computers, data-driven methods have been widely used in various fields. In recent years, some researches about data-driven discovery of partial differential equations (PDEs) have been performed. Among these works, sparse regression and neural network are two main techniques for carrying out PDE discovery.

For data-driven discovery of PDEs using sparse regression, the sparse terms that constitute a PDE are selected from a pre-determined candidate library, which is a collection of potential terms. In existing works, different sparse regression techniques, such as sequential threshold least-squares, sequential threshold ridge regression (STRidge), least absolute shrinkage and selection operator (LASSO), sparse group lasso, and threshold sparse Bayesian regression are utilized to obtain a parsimonious model [2,4,12,13,14,18]. Although sparse regression methods can obtain a parsimonious model with high computational efficiency, these methods may not work well with noisy data, especially if the finite difference method is adopted for calculating the derivatives that are required in the candidate library. Besides the sparse regression technique, neural network is another useful technique that has been employed for data-driven discovery of PDEs. Raissi et al. [9] proposed the physics-informed neural network (PINN) to solve the forward and inverse problem of PDEs. In their work, the PDE terms are supposed to be known, and only the coefficients are learned from data. Long et al. [6] proposed a convolutional neural network-based framework, named PDE-NET, for discovery of PDEs. However, the results may lack parsimony. Xu et al. [17] proposed a deep-learning framework, called DL-PDE, which combines neural network methods and sparse regression methods for PDE discovery. In their work, the neural network is utilized to generate meta-data and calculate derivatives, and sparse regression is then utilized to discover the PDE. Qin et al. [8] and Wu et al. [16] used the residual network (ResNet) to discover underlying PDEs. Hasan et al. [5] introduced a regularization scheme into the loss function of the neural network to prevent overfitting and improve the accuracy of PDE discovery. Compared with numerical methods, neural network can provide a stable calculation of derivatives via automatic differentiation, which makes it more robust to data noise. However, many shortcomings remain for the neural network technique. For example, training the neural network may be time-consuming, and the proper structure of neural networks may be challenging to design.

For both sparse regression and neural network based PDE discovery methods, a pre-defined complete candidate library is usually necessary. This means that the terms that constitute the PDE to be discovered are supposed to be contained in the candidate library. If the candidate library is incomplete, the methods mentioned above will fail to discover the correct PDE. Constructing a complete candidate library is difficult for practical application of PDE discovery methods. A large candidate library is an option, but it cannot be guaranteed to be complete, and it will increase difficulty for sparse identification. Meanwhile, most previous works are constructed on the assumption that the left-hand side of the PDE is the first-order derivative with respect to time. If the left-hand side is not specified with the exact term, these methods may have difficulty in discovering the correct PDE. In order to solve these problems, some recent work combining data-driven methods and an evolutionary approach have been proposed. For example, Maslyaev et al. [7] proposed a data-driven algorithm based on evolutionary optimization to discover PDEs, and compared with STRidge, better performance of the proposed method was achieved. Atkinson et al. [1] presented a framework to discover free-form governing equations with genetic programming algorithms. Compared with traditional data-driven methods, the evolutionary approach can discover PDEs with an incomplete candidate library, which greatly increases its applicability for PDE discovery. However, some drawbacks still exist for evolutionary methods. First, in existing works, derivative calculations are performed using Gaussian process and finite difference methods, which cannot work well with noisy data. Second, evolutionary methods converge slowly and necessitate repeated trials to avoid local minima.

For solving the above issues, in this work, we propose a novel framework for PDE discovery, called DLGA-PDE, which combines the neural network method and genetic algorithm. In our proposed method, a deep neural network is first trained with available data of a physical problem, then utilized to generate meta-data and calculate derivatives, and finally the genetic algorithm is employed to discover the underlying PDE. In the genetic algorithm of DLGA-PDE, a special encoding method is created to digitize the PDE, which is able to handle multiple options for the candidate terms, including the left-hand side (time derivatives). In addition, crossover and mutation are proposed to expand the searching scope of candidate equations, which enables DLGA-PDE to work with an incomplete candidate library. Meanwhile, $l_0$-regularization is utilized in the fitness function to accelerate the rate of calculation and convergence. All of the features mentioned above make DLGA-PDE more flexible and appropriate for dealing with complex problems. Numerical experiments demonstrate that DLGA-PDE works well with limited and noisy data, and converges quickly.

The remainder of this paper is organized as follows. Section 2 describes the methodology of DLGA-PDE in detail. Case studies are presented in Section 3. Discussions and conclusions are given in Section 4.

## 2. Methodology

### 2.1 PDE Discovery

In this work, an extensive form of PDE is investigated as:

$$u_T = \Phi(u) \cdot \zeta \tag{1}$$

with

$$\Phi(u) = [u, u^2, u_x, u_{xx}, ..., uu_x, uu_{xx}, ...] \tag{2}$$

where $u_T$ denotes different orders of derivatives of $u$ with respect to $t$. For example, $u_T$ can be $u_t$ or $u_{tt}$. Different from existing works, with varied options for the temporal derivative, the PDE form shown in Eq. (1) can cover a larger variety of equations. $\Phi(u)$ denotes the candidate library of potential terms in the PDE; and $\zeta$ is the coefficient of corresponding terms.

Given the observation data (or meta-data), which are denoted as $\{u(x_i, t_i)\}_{i=1}^{N}$, we have the

following:

$$\begin{bmatrix} u_T(x_1,t_1) \\ u_T(x_2,t_2) \\ \vdots \\ u_T(x_N,t_N) \end{bmatrix} = \begin{bmatrix} u(x_1,t_1) & \cdots & uu_x(x_1,t_1) & \cdots \\ u(x_2,t_2) & \cdots & uu_x(x_2,t_2) & \cdots \\ \vdots & \ddots & \vdots & \ddots \\ u(x_N,t_N) & \cdots & uu_x(x_N,t_N) & \cdots \end{bmatrix} \cdot \zeta \qquad (3)$$

which can be rewritten as:

$$U_T = \Theta(u) \cdot \zeta \qquad (4)$$

For PDE discovery, it aims to identify the temporal derivative (e.g., whether it is $u_t$ or $u_{tt}$) and the sparse vector $\zeta$ from the data.

*2.2 Architecture of DLGA-PDE*

In this work, we propose a novel framework combining deep learning and genetic algorithm, called DLGA-PDE, for discovering PDEs. Here, we will introduce the architecture of DLGA-PDE. It consists of a neural network step and a genetic algorithm step. Using observation data, a deep neural network is first trained with available data to approximate the response of the considered physical problem. The trained neural network is then utilized to generate meta-data and calculate derivatives. Then, the genetic algorithm is employed to obtain the best model and corresponding coefficients with an incomplete candidate library. In the genetic algorithm step, the PDE is first digitized and encoded to form the corresponding genome. After the procedure of crossover and mutation, fitness of the new generations is calculated to select the best children. Then, the selected children will be the parents of a new generation. This process will continue until convergence. Fig. 1 shows the architecture of DLGA-PDE.

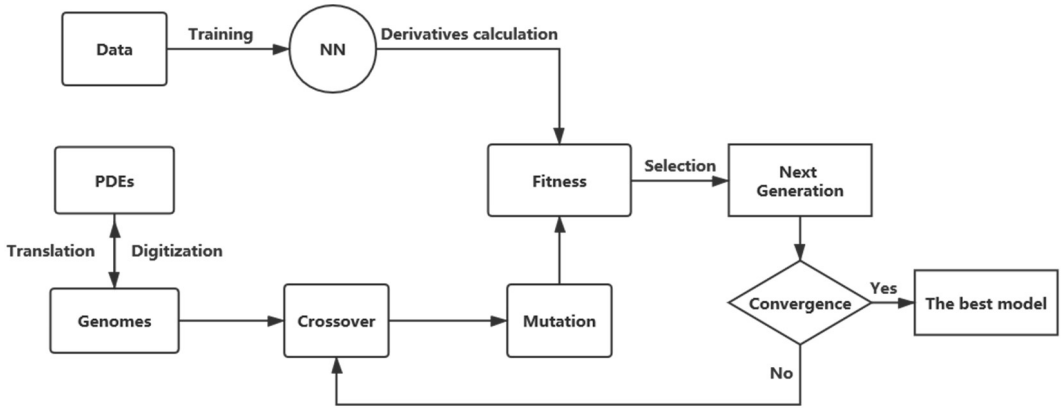

**Fig. 1.** The workflow scheme of DLGA-PDE.

*2.2.1   Neural network*

In the neural network step, a deep feed forward fully-connected artificial neural network (ANN), *NN(x, t)*, is first constructed to approximate *u(x, t)*. Compared with other types of neural networks, the feed forward fully-connected ANN is easier to train and can calculate derivatives relatively more precisely. The structure of a typical ANN is presented in Fig. 2.

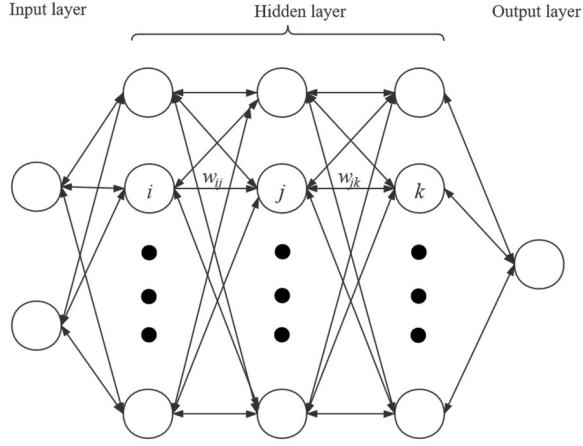

**Fig. 2.** The structure of a typical feed forward fully-connected artificial neural network. *i*, *j*, and *k* are neurons, and $w_{ij}$ and $w_{jk}$ are weights between two neurons.

After designing the neural network structure, training data are utilized to train the neural network by minimizing the loss function:

$$Loss(\theta) = \sum_{i=1}^{N}[u(x_i,t_i) - NN(x_i,t_i;\theta)]^2 \qquad (5)$$

where $\theta$ refers to the parameters to be optimized in the neural network. Meanwhile, certain methods, such as early termination, are used to prevent overfitting. When the training process has completed, the trained neural network will be employed to generate meta-data and calculate derivatives. Here, the procedure of neural network training is only briefly introduced, additional details of which can be found in Xu et al. [17].

*2.2.2    Genetic algorithm*

In this work, a specific genetic algorithm is proposed for discovering PDEs, which will be described in detail below.

(a) Digitization and translation

In order to transform the PDE into a digitized form, the PDE is digitized and encoded to form the corresponding genome.

**Definition 2.1** *Numbers are used to represent the corresponding order of derivatives. For example:*

$$u \leftrightarrow 0, \frac{\partial u}{\partial x} \leftrightarrow 1, \frac{\partial^2 u}{\partial x^2} \leftrightarrow 2, \frac{\partial^3 u}{\partial x^3} \leftrightarrow 3$$

**Definition 2.2** *Each PDE term is considered as a module. Here, it is assumed that there is only multiplication in a module. In fact, most PDEs can be split into a series of multiplication and addition combinations. For example:*

$$u\frac{\partial u}{\partial x} \leftrightarrow [0,1], u\left(\frac{\partial^2 u}{\partial x^2}\right)^2 \leftrightarrow [0,2,2]$$

**Definition 2.3** *Combining these modules yields the genome of the PDE. Modules are connected by a plus sign. It is worth mentioning that there are two parts of the genome component of the PDE. One is the module group of the left-hand side, and the other is the module group of the right-hand side, which is placed in braces. Since only temporal derivatives are in the left-hand side, the term in the left-hand side is encoded in the same way as in Definition 2.1. To simplify the problem, we only consider the case in which the left-hand side has only one term, i.e., the left-hand side has a genome length of 1. In fact, it is easy to extend to a wider range of PDEs. For example, the genome of the contaminant transport equation is:*

$$u_t = -v_x u_x + D_L u_{xx} \leftrightarrow [1], \{[1],[2]\}$$

*Similarly, the genome of the wave equation is:*

$$u_{tt} = A u_{xx} \leftrightarrow [2], \{[2]\}$$

This special encoding method can handle different options for the left-hand side of PDEs, which increases its flexibility. Using the encoding method of PDEs, we can randomly generate a series of genomes in this format, and each genome corresponds to a unique PDE form. In the genetic algorithm, only the basic genes are utilized to generate the first generation. Moreover, the terms in the first generation may contain various combinations of different powers of *u* and different orders of derivatives of *u* with respect to *x*.

(b) Definition of fitness

In the genetic algorithm, fitness refers to the genome's superiority in the population survival measure, which is used to distinguish between good and bad genomes. Fitness is calculated using a fitness function.

**Definition 2.4** *In this algorithm, fitness is defined as:*

$$Fitness = MSE + \varepsilon \cdot len(genome) \qquad (6)$$

where the least square regression is performed on the PDE that is translated from its genome to calculate coefficients; MSE is the mean squared error of the least square regression; $\varepsilon$ is a hyperparameter; and len(genome) is the length of genome. Here, the lower is the fitness, the better is the model.

It is worth mentioning that the fitness function here uses the least square method and $l_0$-regularization, and this sparsity constrained fitness function is not utilized in previous works using the genetic algorithm for PDE discovery [1,7]. Different from sparse regression, the genetic algorithm is not optimized by gradients [15]. In the genetic algorithm, children are produced by crossover and mutation, and are selected according to their fitness. Consequently, it is not optimized along the gradient direction, but rather it is optimized by finding suitable children to evolve. Without the difficulties that it would cause in gradient optimization methods, $l_0$-regularization can be utilized in the genetic algorithm to prevent over-fitting and ensure parsimony. In addition, due to the characteristics of the genetic algorithm, the generated genomes represent the corresponding PDEs, and a large candidate library, such as that in the sparse regression method, is not required.

(c) Process of crossover

For every genome, part of its modules can be replaced with another genome's modules under a certain probability to generate the next generation. For example:

$$Gene1: [1], \{[1],[2]\} \xleftrightarrow{crossover} Gene2: [1], \{[1,3],[0,2]\}$$
$$\downarrow \qquad\qquad\qquad \downarrow$$
$$Gene1': [1], \{[1,3],[2]\} \qquad Gene2': [1], \{[1],[0,2]\}$$

In this work, the crossover rate is 80%, which means that there is an 80% probability of crossover. The process of crossover allows the genes in the parent's genome to be inherited to the next generation, and increases the possibility of different gene combinations.

(d) Process of mutation

Mutation refers to a mutation in some genes in the genome, resulting in a new genome. Here, three ways of mutation are defined.

**Definition 2.5 (Order Mutation)**: *Under a certain probability, the order of derivatives in the gene will be reduced by 1. Particularly, 0 can be mutated to any random higher order. For example:*

$$[1], \{[1,2],[3]]\} \xrightarrow{mutation} [1], \{[0,2],[3]\}$$

$$[1], \{[1,3],[0]\} \xrightarrow{mutation} [1], \{[1,3],[2]\}$$

**Definition 2.6 (Add-module Mutation)**: *Under a certain probability, a random module is added to the genome. For example:*

$$[1],\{[1,2],[3]\} \xrightarrow{mutation} [1],\{[1,2],[3],[0,0]\}$$

**Definition 2.7 (Delete-module Mutation)**: *Under a certain probability, a module is deleted from the genome. For example:*

$$[1],\{[1,2],[4],[0,1],[3,1]\} \xrightarrow{mutation} [1],\{[1,2],[4],[0,1]\}$$

Different from previous works using the genetic algorithm for PDE discovery, here, a variety of mutation methods are set. Order Mutation can change the order of the derivatives in the genome. On the other hand, Add-module Mutation and Delete-module Mutation can alter the genome length. These three different mutation ways may also occur independently. The diversification of mutation methods enables the genome to possibly have diverse changes, which is helpful for identifying the best model.

It is worth noting that there are three mutation ways for the right-hand side of the equation, but the left-hand side of the equation only has Order Mutation. This means that the left-hand side can only change the derivative order and its length remains the same, i.e., only one term.

(e) Process of selection and evolution

In the process of crossover, each parent genome crossovers twice, producing twice as many genomes as the parent genome. For example, 50 parents will produce 100 children via crossover. Then, the children's fitness will be calculated and sorted from small to large. Finally, the first half of children will be selected as a new generation of parents. We set the number of genomes and the number of generations. After certain epochs, the genome that occurs stably during evolution is the best model.

3. **Results**

In this section, some classical PDEs will be utilized to test the performance of DLGA-PDE.

*3.1 Problem Setting*

In this part, we will briefly introduce the settings of the discussed PDEs, which are the Korteweg–de Vries (KdV) equation, the wave equation, the Chaffee-Infante equation, and the Burgers equation. Additional details about the generation of the dataset and meta-data are provided in Supplementary Information, Section S1.

*3.1.1  KdV equation*

The KdV equation takes the following form:

$$u_t = -uu_x + bu_{xxx} \tag{7}$$

The conventional spectral method is utilized to generate the dataset [9]. In this example, we set $b=0.0025$. In the dataset, there are 201 temporal observation steps at intervals of 0.005, and 512 spatial observation steps at intervals of 0.0039. Therefore, the total number of data points is 102912. A five-layer deep neural network with 50 neurons per hidden layer is constructed. The activation function is sin(*x*). 30000 data from the dataset are randomly selected to train the neural network. Then, the trained neural network is utilized to generate 200000 meta-data, and derivatives are calculated using automatic differentiation.

*3.1.2 Wave equation*

The wave equation takes the form:

$$u_{tt} = Au_{xx} \tag{8}$$

The central difference method is utilized to generate the dataset. In this work, we set *A*=1. In the dataset, there are 321 temporal observation steps at intervals of 0.0196, and 161 spatial observation steps at intervals of 0.0196. Therefore, the total number of data points is 51681. A five-layer deep neural network with 50 neurons per hidden layer is constructed. The activation function is sin(*x*). 10000 data are randomly selected to train the neural network. Then, the trained neural network is utilized to generate 160000 meta-data, and derivatives are calculated using automatic differentiation.

*3.1.3 Burgers equation*

The Burgers equation takes the following form:

$$u_t = -uu_x + au_{xx} \tag{9}$$

The conventional spectral method is utilized to generate the dataset. In this case, we set *a*=1. In the dataset, there are 201 temporal observation steps at intervals of 0.05, and 256 spatial observation steps at intervals of 0.0625. Therefore, the total number of data points is 51456. A nine-layer deep neural network with 20 neurons per hidden layer is constructed. The activation function is tanh(*x*). 2000 data are randomly selected to train the neural network. Then, the trained neural network is utilized to generate 57600 meta-data, and derivatives are calculated using automatic differentiation.

*3.1.4 Chaffee-Infante equation*

The Chaffee-Infante equation takes the form:

$$u_t = u_{xx} + \lambda(u^3 - u) \tag{10}$$

In Eq. (10), $\lambda > 0$ is the diffusion coefficient. When $\lambda = 1$, Eq. (10) is also called the Whitehead equation. It is widely used in many fields, such as environmental science, fluid dynamics, high-energy physics, and electronic science. In this work, we set $\lambda = 1$.

In this work, the forward difference method is utilized to generate the dataset. In the dataset, there are 200 temporal observation steps at intervals of 0.002, and 301 spatial observation steps at intervals of 0.01. Therefore, the total number of data points is 60200. A five-layer deep neural network with 50 neurons per hidden layer is constructed. The activation function is sin($x$). 10000 data are randomly selected to train the neural network. Then, the trained neural network is utilized to generate 160000 meta-data, and derivatives are calculated using automatic differentiation.

*3.2 Settings and Performance of Standard DLGA-PDE*

For DLGA-PDE, a standard setting is first utilized, which can handle most PDE discovery problems. In the standard setting, basic genes are set to be $[u_t(1), u_{tt}(2)]\{u(0), u_x(1), u_{xx}(2), u_{xxx}(3)\}$, and thus the first generation is randomly generated by a combination of these genes. Here, we consider up to the third-order derivative for the right-hand side terms, which means that the highest-order derivative produced by mutation is third-order in the right-hand side term. In fact, most PDEs are composed of derivatives of the third-order and below. Therefore, it is reasonable to only consider the third-order derivative for the standard setting. The size of the population is set to be 200, and the maximum number of generations is set to be 100.

At first, the four PDEs mentioned above are utilized to test the performance of standard DLGA-PDE. The results are shown in Table 1. It can be seen that standard DLGA-PDE has successfully discovered true PDE terms, which shows that DLGA-PDE is able to handle various types of PDEs. At the same time, the estimations of the coefficients of the PDE terms are very accurate. It is worth mentioning that the candidate library utilized here is incomplete, but DLGA-PDE can still discover the correct PDE owing to the characteristics of the genetic algorithm.

**Table 1.** Summary of the learned structure via standard DLGA-PDE for four examples.

|  | Discovered Structure | True Structure |
|---|---|---|
| **KdV Equation** | $u_t, u_{xxx}, uu_x$ | $u_t, u_{xxx}, uu_x$ |
| **Wave Equation** | $u_{tt}, u_{xx}$ | $u_{tt}, u_{xx}$ |
| **Burgers Equation** | $u_t, u_{xx}, uu_x$ | $u_t, u_{xx}, uu_x$ |
| **Chaffee-Infante Equation** | $u_t, u_{xx}, u, u^3$ | $u_t, u_{xx}, u, u^3$ |

In order to illustrate the procedure of DLGA-PDE more explicitly, Table 2 shows the best child found in each generation when DLGA-PDE is utilized to discover the Chaffee-Infante equation.

**Table 2.** Best child found in each generation when DLGA-PDE is utilized to discover the Chaffee-Infante equation.

| Generation | Structure of Discovered Best Child |
|:---:|:---:|
| 1 | $u_t, u^2, u_x, u_{xxx}, uu_{xx}, u^2u_{xx}$ |
| 2 | $u_t, u, u_x^2, uu_{xx}, uu_x, u_{xx}u_{xxx}$ |
| 4 | $u_t, u_{xx}, uu_{xxx}, u_{xx}^2, u_{xxx}$ |
| 5 | $u_t, u_{xx}, uu_{xxx}, u_{xx}^2, uu_{xx}u_{xxx}, u$ |
| 6 | $u_t, u_{xx}, u^2, u^2u_{xx}, u^3$ |
| 36 | $u_t, uu_{xxx}, u^2, u, u^3$ |
| 39 | $u_t, u_{xx}, u^2, uu_{xx}^2u_{xxx}, u^3$ |
| 40 | $u_t, u_{xx}, u, u^3$ |
| 100 | $u_t, u_{xx}, u, u^3$ |

Due to the difficulty of this example, it took many generations to converge and discover the correct structure. For this case, in the first several generations (e.g., 1-5[th] generation), the best child contained many terms and fell into a local minimum in the sixth generation. Furthermore, in the 39[th] generation, it jumped out of the local minimum and continued to evolve until it found the best model. From this case, one can see that the process of mutation and crossover in DLGA-PDE can effectively prevent becoming stuck with the local minimum. It is worth noting that $u^3$ has appeared in the sixth generation, which demonstrates that DLGA-PDE has the ability to produce more complex combinations, and it can search far beyond the incomplete candidate library compared with traditional methods.

*3.3 Effect of Mutation and Combination*

In order to further investigate the effect of mutation and combination in the genetic algorithm step of DLGA-PDE, the basic genes will be changed in the next few numerical experiments. The performance of DLGA-PDE will be assessed when some basic order of derivatives is not contained in the basic genes.

*3.3.1 Cases of missing high-order derivatives*

In order to test the effect of mutation proposed in DLGA-PDE, DLGA-PDE is utilized to discover the wave equation and the KdV equation again in the absence of some order of derivatives in basic genes. Specially, the missing derivative terms in the basic genes are parts of the target PDE, and thus it is impossible to form the target PDE based on the provided basic genes without mutation.

The KdV equation is first investigated, which has a third-order derivative term. Different from the previous standard settings, here the basic genes are changed to be $[u_t(1), u_{tt}(2)]\{u(0), u_x(1), u_{xx}(2)\}$. The third-order derivative term is not included in the basic genes. In order to increase the difficulty for obtaining the correct term (the third-order derivative term), we set that the highest-order derivative produced by mutation is fourth-order for the right-hand side terms. Similarly, the best child found in each generation is displayed in Table 3.

**Table 3.** Best child found in each generation when DLGA-PDE is utilized to discover the KdV equation in the absence of a third-order derivative term in basic genes.

| Generations | Discovered Structure |
|---|---|
| 1 | $u_t, u_{xx}, uu_x, u_x u_{xx}, u$ |
| 2 | $u_t, u_{xxx}, u_x, uu_x$ |
| 3 | $u_t, u_{xxx}, uu_x$ |
| 100 | $u_t, u_{xxx}, uu_x$ |

From Table 3, it can be seen that the best child in the first generation does not have new higher-order derivatives. In the second generation, however, the third-order derivative is generated by mutation. The best child remains unchanged from the third generation to the last generation, which means that the algorithm has converged. Finally, the correct form of PDE is discovered in the absence of the third-order derivative term in basic genes. This shows that DLGA-PDE can effectively find higher-order derivative terms via mutation, even if they are not included in the basic genes. In addition, DLGA-PDE exhibits a fast convergence rate and good stability.

The wave equation is then investigated, whose left-hand side term is $u_{tt}$. Previous experiments have demonstrated that DLGA-PDE can discover the correct left-hand side term for the wave equation with standard settings. Here, the basic genes are set as $[u_t(1)]\{u(0), u_x(1), u_{xx}(2), u_{xxx}(3)\}$. We can see that the left-hand side term only has $u_t$ in basic genes, which means that $u_{tt}$ can only be produced by mutation. Other settings are the same as in the previous case. The best child found in each generation is presented in Table 4. From Table 4, it can be seen that $u_{tt}$ is obtained in the best child in the second generation by mutation. Meanwhile, the correct PDE form is discovered very quickly.

**Table 4.** Best child found in each generation when DLGA-PDE is utilized to discover the wave equation in the absence of $u_{tt}$ in basic genes.

| Generations | Discovered Structure |
|---|---|
| 1 | $u_t, u^2$ |
| 2 | $u_{tt}, u_{xx}$ |
| 100 | $u_{tt}, u_{xx}$ |

In the above, two cases with missing high-order derivative terms in basic genes are utilized to test the effect of mutation. In these situations, DLGA-PDE successfully produces these missing terms via mutation. The missing term can be a high-order derivative with respect to *x* or a high-order derivative with respect to *t*. It can be found that DLGA-PDE can still quickly converge because of the effect of mutation in these cases. This also demonstrates that DLGA-PDE can adapt to many special situations with low restriction of the candidate library.

*3.3.2 Case of missing terms that need mutation and combination*

Next, a more difficult situation with missing terms in the basic genes is investigated. For obtaining the missing terms, both mutation and combination are required in the genetic algorithm step. Here, Burgers is investigated again. Different from standard settings, the basic genes are changed to be $[u_t(1), u_{tt}(2)]\{u(0), u_{xx}(2)\}$, and up to the third-order derivatives are still considered here.

Because $u_x$ is not included in basic genes, $uu_x$ must be produced by both mutation and combination for discovering the correct Burgers equation. The best child found in each generation is displayed in Table 5. From Table 5, one can see that DLGA-PDE converges very quickly and successfully finds the correct structure. From this case, it can be seen that DLGA-PDE can effectively discover terms that need both mutation and combination.

**Table 5.** Best child found in each generation when DLGA-PDE is utilized to discover the Burgers equation in the absence of a necessary term in basic genes.

| Generations | Discovered Structure |
|---|---|
| 1 | $u_t, u^3$ |
| 2 | $u_t, u_{xx}, uu_x$ |
| 50 | $u_t, u_{xx}, uu_x$ |

*3.4 Comparison with DL-PDE*

In order to further test the performance of DLGA-PDE, it is compared with DL-PDE, which utilizes the neural network to calculate derivatives and utilizes sparse regression to discover PDEs. Additional details about DL-PDE can be found in Xu et al. [17]. The major difference between DLGA-PDE and DL-PDE is that DL-PDE needs a complete candidate library, while DLGA-PDE can work with an incomplete candidate library. The Burgers equation and the Chaffee-Infante equation are utilized to test the performance of these two methods.

For the Burgers equation, different levels of noise are added to data, and 1000 data (1.94% of total data) are randomly selected to train the neural network. In this work, noise is added as follows:

$$u(x,t) = u(x,t) \times (1 + \delta \times e) \qquad (11)$$

where $\delta$ denotes the noise level; and $e$ is the uniform random variable, taking values from -1 to 1 [3].

In this example, the performance of the two methods will be studied on a small amount of data and different noise levels. In this case, DLGA-PDE uses the standard settings, as discussed previously. In addition, DL-PDE uses a candidate library as follows:

$$\Phi = [1 \quad u \quad u^2 \quad u_x \quad uu_x \quad u^2 u_x \quad u_{xx} \quad uu_{xx} \quad ... \quad uu_{xxx} \quad u^2 u_{xxx}] \qquad (12)$$

Then, DLGA-PDE and DL-PDE are utilized to discover the Burgers equation with different noise levels, and the results are provided in Table 6. From Table 6, it can be seen that both DLGA-PDE and DL-PDE perform well when data are limited and noisy, but DLGA-PDE is more stable when the noise level is 20%.

**Table 6.** For discovering the Burgers equation, the results of DL-PDE and DLGA-PDE with different noise levels.

| Noise Level | Learned Equation (1000 Data) | |
|---|---|---|
| | Learned Equation (DL-PDE) | Learned Equation (DLGA-PDE) |
| Correct PDE | $u_t = -uu_x + 0.1u_{xx}$ | $u_t = -uu_x + 0.1u_{xx}$ |
| Clean Data | $u_t = -0.957uu_x + 0.088u_{xx}$ | $u_t = -0.957uu_x + 0.088u_{xx}$ |
| 1% Noise | $u_t = -0.953uu_x + 0.087u_{xx}$ | $u_t = -0.953uu_x + 0.087u_{xx}$ |
| 5% Noise | $u_t = -0.886uu_x + 0.079u_{xx}$ | $u_t = -0.886uu_x + 0.079u_{xx}$ |
| 10% Noise | $u_t = -0.824uu_x + 0.077u_{xx}$ | $u_t = -0.824uu_x + 0.077u_{xx}$ |
| 15% Noise | $u_t = -0.637uu_x + 0.059u_{xx}$ | $u_t = -0.637uu_x + 0.059u_{xx}$ |
| 20% Noise | $u_t = -0.032u_x - 0.585u_{xx} + 0.057u^2$ | $u_t = -0.570uu_x + 0.055u_{xx}$ |

Using the Chaffee-Infante equation, we will study the performance of the two methods on different amounts of data with other conditions being identical. Similarly, DLGA-PDE uses the standard settings, while DL-PDE uses a candidate library with 16 terms:

$$\Phi = [1 \quad u \quad u^2 \quad u^3 \quad u_x \quad uu_x \quad u^2 u_x \quad u^3 u_x \quad ... \quad u^2 u_{xxx} \quad u^3 u_{xxx}] \quad (13)$$

Then, DLGA-PDE and DL-PDE are utilized to discover the Chaffee-Infante equation with different amounts of data, the results of which are presented in Table 7. From Table 7, it can be seen that DLGA-PDE performs well with limited data, while DL-PDE fails in all experiments, which indicates that DLGA-PDE is much more stable than DL-PDE in this case.

From the two experiments above, it can be found that the accuracy of DL-PDE is affected by the size of the candidate library. For the Burgers equation, only 12 terms are included in the candidate library. On the other hand, for the Chaffee-Infante equation, due to the source and sink term, which contains $u^3$ and $u$, many terms must be included in the candidate library. Consequently, the candidate library for discovering the Chaffee-Infante equation is larger than that for the Burgers equation, which may be the reason for the failure to discover the correct Chaffee-Infante equation. For a larger candidate library, greater sparsity will be required when performing the sparse regression that is utilized in DL-PDE, which may increase the difficulty. In contrast,

DLGA-PDE can discover PDEs with an incomplete candidate library, and the PDE terms to be discovered can be produced by mutation and crossover of genomes. In this way, the problems with the candidate library that occur in DL-PDE can be avoided in DLGA-PDE. By comparison with DL-PDE, the advantages of DLGA-PDE can be clearly discerned.

**Table 7.** For discovering the Chaffee-Infante equation, the results of DL-PDE and DLGA-PDE with different data volume.

| Volume of Data | Learned Equation (Clean Data) | |
| --- | --- | --- |
| | **Learned Equation (DL-PDE)** | **Learned Equation (DLGA-PDE)** |
| **Correct PDE** | $u_t = u_{xx} + u^3 - u$ | $u_t = u_{xx} + u^3 - u$ |
| **10000 Data** (16.6% of Total) | $u_t = 0.636 u_{xx} + 0.611 u^3$ | $u_t = 0.999 u_{xx} + 1.000 u^3 - 1.001 u$ |
| **2500 Data** (4.15% of Total) | $u_t = 0.709 u^2$ | $u_t = 1.009 u_{xx} + 1.004 u^3 - 1.034 u$ |
| **1000 Data** (1.66% of Total) | $u_t = -1.31 u^2 + 0.76 u_{xx} + 1.42 u^3 + 0.13 u u_{xx}$ | $u_t = 1.027 u_{xx} + 1.034 u^3 - 1.090 u$ |
| **500 Data** (0.83% of Total) | $u_t = 0.865 u_{xx} + 1.133 u^3 - 0.896 u^2$ | $u_t = 0.959 u_{xx} + 0.953 u^3 - 0.893 u$ |

*3.5 DLGA-PDE with Limited and Noisy Data*

Using the Burgers equation and the Chaffee-Infante equation, it can be found that DLGA-PDE performs well when the data are limited and noisy. In order to further demonstrate the robustness of DLGA-PDE to limited and noisy data, DLGA-PDE is utilized to discover more PDEs with limited and noisy data, additional details of which are provided in the Supplementary Information, Section S2. It is shown that DLGA-PDE can find the correct PDE even if the noise level is 15% and performs well with little data, which means that DLGA-PDE is able to handle a large variety of PDEs with noisy and limited data.

## 4. Conclusion and Discussion

In this work, we propose a novel framework, called DLGA-PDE, for discovering underlying PDEs with an incomplete candidate library, which combines the neural network method and the genetic algorithm. In the proposed method, a new way to digitize and code the PDE is proposed, and special principles of mutation are defined, which makes it able to quickly converge and avoid falling into local minima. Compared with ordinary genetic methods, DLGA-PDE is robust to limited and noisy data. The most significant advantage of DLGA-PDE is that it can discover PDEs with an incomplete candidate library. Through crossover and mutation, infinite possible genomes may be created, which greatly expands the search scope of DLGA-PDE and makes it more flexible for PDE discovery. Owing to the merit of its unique encoding method, DLGA-PDE can find the correct PDE when the left-hand side term of PDE has multiple options (e.g., $u_t$, $u_{tt}$, and so on), which may not be handled well in traditional methods. In DLGA-PDE, $l_0$-regularization is adopted to construct the fitness function, and by using the $l_0$-regularization, it increases the stability and efficiency.

In this work, different numerical experiments are investigated, and the results demonstrate that DLGA-PDE performs well for a wide range of PDEs with standard settings. The effect of mutation and combination of DLGA-PDE is examined by setting different basic genes, and the results show that DLGA-PDE can find the correct PDE term via mutation and combination of genomes, even if some basic order of derivatives are not included in the basic genes. This demonstrates that DLGA-PDE can work well when scarce information is known about the PDE structure. By recording the evolution process, one can see that DLGA-PDE converges quickly in most cases. Meanwhile, DLGA-PDE is robust to limited and noisy data. This is because derivatives are calculated by automatic differentiation of the trained neural network with a large number of meta-data.

By comparing the performance of DL-PDE and DLGA-PDE, the results of these two methods do not show much difference when the candidate library of DL-PDE is relatively small. However, for the Chaffee-Infante equation, the DL-PDE method fails to find the correct PDE form because a large candidate library is utilized to account for more complex terms. In contrast, DLGA-PDE works well and finds the correct terms stably for discovering the Chaffee-Infante equation. This demonstrates that DLGA-PDE has better stability and wider applicability in practical applications.

Some limitations and shortcomings still exist for DLGA-PDE. The hyper-parameter of $l_0$-regularization is important for discovering the correct PDE form. However, it is currently set by experience, and a sophisticated method for adjusting the hyper-parameter is needed. In addition, DLGA-PDE can only be utilized under the condition that coefficients of the PDE are constant in space and time. It cannot now deal with certain other problems, such as piecewise-constant coefficients [3] or smoothly varying coefficients [11]. Further works regarding this issue are necessary.

**Acknowledgements**

This work is partially funded by the National Natural Science Foundation of China under Grants 51520105005 and U1663208, and the National Science and Technology Major Project of China (Grant No. 2017ZX05009-005 and 2017ZX05049-003).

# Supplementary Information

**DLGA-PDE: Discovery of PDEs with incomplete candidate library via combination of deep learning and genetic algorithm**

Hao Xu[a], Haibin Chang[a,*], and Dongxiao Zhang[b,c,*]

[a] BIC-ESAT, ERE, and SKLTCS, College of Engineering, Peking University, Beijing 100871, P. R. China

[b] School of Environmental Science and Engineering, Southern University of Science and Technology, Shenzhen 518055, P. R. China

[c] Intelligent Energy Lab, Frontier Research Center, Peng Cheng Laboratory, Shenzhen 518055, P.R. China

* Corresponding author.

E-mail address: 390260267@pku.edu.cn (H. Xu); changhaibin@pku.edu.cn (H. Chang); donzhang01@gmail.com (D. Zhang).

## Section S1. Additional details about the discussed PDEs

### 1.1 KdV equation

The Korteweg–de Vries (KdV) equation is a partial differential equation describing one-way motion of shallow water waves. It was discovered by Korteweg and de Vries when investigating small-amplitude and long-wave motion in shallow water. Its equation takes the form:

$$u_t = -uu_x - bu_{xxx} \tag{S1}$$

where $b$ is a constant.

To obtain a dataset, the conventional spectral method is utilized to solve the KdV equation. We start with an initial condition being $u(0,x) = \cos(\pi x)$, $x \in [-1,1]$, and periodic boundary conditions. Eq. (S1) is integrated from the starting time $t=0$ to the final time $t=1$, which is done by utilizing the Chebfun package with a spectral Fourier discretization with 512 modes and a fourth-order explicit Runge-Kutta temporal integrator with time-step size being $10^{-6}$ [9]. The solution is recorded every $\Delta t = 0.005$ to obtain 201 observation steps in time. Therefore, we have $n_x=512$, $n_t=201$, and $N_d=102912$. To generate meta-data, we take spatial observation steps with $\Delta x = 0.001$ in the domain $x \in [-0.5, 0.5)$, and take temporal observation steps with $\Delta t = 0.005$ in the domain $t \in [0,1)$. Thus, for meta-data, we have $n'_x$=1000, $n'_t$=200, and $N'_d$=200000.

**1.2 Wave equation**

The wave equation is an important partial differential equation, which mainly describes various wave phenomena in nature, including transverse and longitudinal waves, such as acoustic, light, and water waves. It arises in numerous fields, such as acoustics, electromagnetics, and fluid mechanics. Its equation takes the following form:

$$u_{tt} = A u_{xx} \tag{S2}$$

To obtain a dataset, the central differential method is utilized to solve the wave equation. Three-point central difference is used to approximate the second-order derivative, and this equation is discretized as:

$$\frac{u_{i,j+1} - 2u_{i,j} + u_{i,j-1}}{\Delta t^2} = A \frac{u_{i+1,j} - 2u_{i,j} + u_{i-1,j}}{\Delta x^2} \tag{S3}$$

where $\Delta x = \frac{x_f}{M}$; $\Delta t = \frac{T}{N}$; $x_f$ is the right end of the spatial domain; $T$ is the right end of the temporal domain; $M$ is the number of nodes in the $x$ direction; and $N$ is the number of nodes in the $t$ direction. By defining $r = A \frac{\Delta t^2}{\Delta x^2}$, Eq. (S3) is simplified as:

$$u_{i,j+1} = r(u_{i-1,j} + u_{i+1,j}) + (2-2r)u_{i,j} - u_{i,j-1} \tag{S4}$$

In this case, initial conditions are as follows:

$$u(x,0) = \begin{cases} \frac{\sin(2x)}{2}, & 0 \le x < \frac{\pi}{2} \\ 0, & \frac{\pi}{2} \le x \le \pi \end{cases}, \quad \frac{\partial u}{\partial t}(x,0) = 0 \tag{S5}$$

Meanwhile, boundary conditions are set to be: $u(0,t) = u(1,t) = 0, t > 0$. Other parameters are set as: $A$=1; $x_f = \pi$; $M$=160; $T = 2\pi$; and $N$=320. Matlab is utilized to solve the problems for obtaining the dataset. There are 321 temporal observation steps at intervals of 0.0196 and 161 spatial observation steps at intervals of 0.0196, and thus the total number of data points is 51681. To generate meta-data, we take 400 spatial observation steps uniformly in the domain $x \in [0,2]$, and take 400 temporal observation steps uniformly in the domain $t \in [0,5]$. Therefore, for meta-data, we have $n'_x=400$, $n'_t=400$, and $N'_d=160000$.

**1. 3 Burgers equation**

The Burgers equation is a nonlinear partial differential equation that simulates the propagation and reflection of shock waves. The Burgers equation is the basic partial differential equation for various fields of applied mathematics, such as fluid mechanics, nonlinear acoustics, and gas dynamics. One-dimensional Burgers equation can take the following form:

$$u_t = -uu_x + au_{xx} \qquad (S6)$$

where $a$ is the diffusion coefficient, which is set to be 0.1 in this example. To obtain a dataset, the conventional spectral method is utilized to solve the Burgers equation with an initial condition being $u(0,x) = -\sin(\pi x/8)$, $x \in [-8,8]$, and periodic boundary conditions. Eq. (S6) is integrated from the starting time $t=0$ to the final time $t=10$, which is accomplished by using the Chebfun package with a spectral Fourier discretization with 256 modes and a fourth-order explicit Runge-Kutta temporal integrator with time-step size being $10^{-4}$ [10]. The solutions are recorded every $\Delta t = 0.05$ to obtain 201 observation steps in time. Thus, we have $n_x=256$, $n_t=201$, and $N_d=51456$. To generate meta-data, we take spatial observation steps with $\Delta x = 0.05$ in the domain $x \in [-8,8)$, and take temporal observation steps with $\Delta t = 0.05$ in the domain $t \in [0,9)$. Therefore, for these meta-data, we have $n'_x=320$, $n'_t=180$, and $N'_d=57600$.

**1.4 Chaffee-Infante equation**

The Chaffee-Infante equation is an important nonlinear partial differential equation, which takes the form:

$$u_t = u_{xx} + \lambda(u^3 - u) \qquad (S7)$$

In Eq. (S7), $\lambda > 0$ is the diffusion coefficient. When $\lambda = 1$, Eq. (S7) is also called the Whitehead equation. It is widely used in numerous fields, such as environmental science, fluid dynamics, high-energy physics, and electronic science. In this work, we set $\lambda = 1$. To obtain a dataset, the explicit forward Euler method is utilized to solve the wave equation. Thus, it can be

discretized as follows:

$$\frac{u_{i,j+1} - u_{i,j}}{\Delta t} = \frac{u_{i-1,j} - 2u_{i,j} + u_{i+1,j}}{\Delta x^2} + \lambda(u_{i,j}^3 - u_{i,j}) \quad (S8)$$

where $\Delta x = \frac{x_f}{M}$; $\Delta t = \frac{T}{N}$; $x_f$ is the right end of the spatial domain; $T$ is the right end of the temporal domain; $M$ is the number of nodes in the $x$ direction; and $N$ is the number of nodes in the $t$ direction.

By defining $r = \frac{\Delta t}{\Delta x^2}$, Eq. (S8) can be simplified as:

$$u_{i,j+1} = (1-2r)u_{i,j} + r(u_{i-1,j} + u_{i+1,j}) + \lambda(u_{i,j}^3 - u_{i,j})\Delta t \quad (S9)$$

In this case, initial conditions are set to be: $u(x,0) = x\sin(x), 0 \leq x \leq 3$. Meanwhile, boundary conditions are set to be: $u(0,t) = u(3,t) = 0, t > 0$. Other parameters are set as: $\lambda = 1$; $x_f = 3$; $M=300$; $T = 0.5$; and $N=80000$. Matlab is utilized to solve this problem. To simplify the dataset, one temporal observation step is taken from every 320 numerical time steps from $t=0.1$ to $t=0.5$, and thus there are a total of 200 temporal observation steps. In addition, there are 301 spatial observation steps at intervals of 0.01. Therefore, the total number of data points is 60200. To generate meta-data, we take 400 spatial observation steps uniformly in the domain $x \in [0.3, 2]$, and take 400 temporal observation steps uniformly in the domain $t \in [0.2, 0.4]$. Thus, for meta-data, we have $n'_x = 400$, $n'_t = 400$, and $N'_d = 160000$.

## Section S2. Results of DLGA-PDE with limited and noisy data

### 2.1 Discovery of KdV equation with noisy data

For the KdV equation, different levels of noise are added to the data, and 30000 data are randomly selected to train the neural network. DLGA-PDE uses the standard setting, and the results are shown in Table S1. From Table S1, it can be seen that DLGA-PDE performs well when data are noisy.

**Table S1.** Summary of the learned equation for discovering the KdV equation using DLGA-PDE with noisy data.

| Noise Level | Learned Equation |
|---|---|
| Correct PDE | $u_t = -uu_x - 0.0025 u_{xxx}$ |
| Clean Data | $u_t = -0.993 uu_x - 0.00248 u_{xxx}$ |
| 1% Noise | $u_t = -0.990 uu_x - 0.00247 u_{xxx}$ |
| 5% Noise | $u_t = -0.973 uu_x - 0.00243 u_{xxx}$ |
| 10% Noise | $u_t = -0.952 uu_x - 0.00236 u_{xxx}$ |
| 15% Noise | $u_t = -0.887 uu_x - 0.0022 u_{xxx}$ |

## 2.2 Discovery of wave equation with noisy and limited data

For the wave equation, different amounts of data are utilized to train the neural network. DLGA-PDE uses the standard setting, and the results are presented in Table S2.

**Table S2.** Summary of the learned equation for discovering the wave equation using DLGA-PDE with different amounts of data training the neural network.

| Volume of Data | Learned Equation |
|---|---|
| Correct PDE | $u_{tt} = u_{xx}$ |
| 10000 Data (19.35% of Total) | $u_{tt} = 0.9998 u_{xx}$ |
| 5000 Data (9.67% of Total) | $u_{tt} = 0.9867 u_{xx}$ |
| 2000 Data (3.87% of Total) | $u_{tt} = 0.9752 u_{xx}$ |
| 500 Data (0.967% of Total) | $u_{tt} = 0.9532 u_{xx}$ |

| 300 Data (0.580% of Total) | $u_{tt} = 0.892 u_{xx}$ |
|---|---|
| 100 Data (0.193% of Total) | $u_{tt} = 0.604 u_{xx} + 0.108 u_{xx}^2$ |

Next, different levels of noise are added to the original data, and 2000 data are selected to train the neural network. The results are shown in Table S3. From Table S2 and Table S3, it can be seen that DLGA-PDE performs well with limited and noisy data.

**Table S3.** Summary of the learned equation for discovering the wave equation using DLGA-PDE with noisy data.

| Noise Level | Learned Equation |
|---|---|
| Correct PDE | $u_{tt} = u_{xx}$ |
| Clean Data | $u_{tt} = 0.9752 u_{xx}$ |
| 1% Noise | $u_{tt} = 0.991 u_{xx}$ |
| 5% Noise | $u_{tt} = 0.960 u_{xx}$ |
| 10% Noise | $u_{tt} = 0.942 u_{xx}$ |
| 15% Noise | $u_{tt} = 0.938 u_{xx}$ |
| 20% Noise | $u_{tt} = -1.662 u^2$ |

## 2.3 Discovery of Chaffee-Infante equation with noisy data

For the Chaffee-Infante equation, different levels of noise are added to the data, and 10000 data are randomly selected to train the neural network. DLGA-PDE uses the standard setting, and the results are presented in Table S4. From Table S4, one can see that DLGA-PDE performs well when the data are noisy.

**Table S4.** Summary of the learned equation for discovering the Chaffee-Infante equation using DLGA-PDE with noisy data.

| Noise Level | Learned Equation |
|---|---|
| **Correct PDE** | $u_t = u_{xx} + u^3 - u$ |
| **Clean Data** | $u_t = 0.999 u_{xx} + 1.000 u^3 - 1.001 u$ |
| **1% Noise** | $u_t = 0.996 u_{xx} + 1.001 u^3 - 1.007 u$ |
| **5% Noise** | $u_t = 1.010 u_{xx} + 0.993 u^3 - 0.973 u$ |
| **10% Noise** | $u_t = 1.032 u_{xx} + 1.021 u^3 - 1.072 u$ |
| **15%** | $u_t = 1.010 u_{xx} + 0.993 u^3 - 0.973 u$ |